\documentclass[runningheads]{llncs}
\usepackage{graphicx}
\usepackage{booktabs}
\usepackage{amsmath,amssymb}
\usepackage{color}
\usepackage[dvipsnames]{xcolor}
\usepackage{bm}
\usepackage{bbm}
\usepackage[caption=false]{subfig}
\usepackage{hyperref}
\hypersetup{
    colorlinks=true,
    urlcolor=purple,
    }

\newcommand{\etal}{\textit{et al}.}

\newcommand{\eg}{\textit{e}.\textit{g}.}

\newcommand{\tinytit}[1]{\noindent\textbf{#1.}}

\begin{document}
%
\title{Out of the Box:\\Embodied Navigation in the Real World}
%
%
\author{
%
Roberto Bigazzi \and
Federico Landi \and
Marcella Cornia \\
Silvia Cascianelli \and 
Lorenzo Baraldi \and 
Rita Cucchiara
}

\authorrunning{R. Bigazzi et al.}
%
\institute{University of Modena and Reggio Emilia, Modena, Italy\\
\email{\{name.surname\}@unimore.it}}
\maketitle              
\begin{abstract}
The research field of Embodied AI has witnessed substantial progress in visual navigation and exploration thanks to powerful simulating platforms and the availability of 3D data of indoor and photorealistic environments. These two factors have opened the doors to a new generation of intelligent agents capable of achieving nearly perfect PointGoal Navigation. However, such architectures are commonly trained with millions, if not billions, of frames and tested in simulation. Together with great enthusiasm, these results yield a question: how many researchers will effectively benefit from these advances?
In this work, we detail how to transfer the knowledge acquired in simulation into the real world. To that end, we describe the architectural discrepancies that damage the Sim2Real adaptation ability of models trained on the Habitat simulator and propose a novel solution tailored towards the deployment in real-world scenarios. We then deploy our models on a LoCoBot, a Low-Cost Robot equipped with a single Intel RealSense camera.
Different from previous work, our testing scene is unavailable to the agent in simulation. The environment is also inaccessible to the agent beforehand, so it cannot count on scene-specific semantic priors. In this way, we reproduce a setting in which a research group (potentially from other fields) needs to employ the agent visual navigation capabilities as-a-Service. Our experiments indicate that it is possible to achieve satisfying results when deploying the obtained model in the real world.
Our code and models are available at \url{https://github.com/aimagelab/LoCoNav}.
\end{abstract}
\section{Introduction}
Embodied AI has recently attracted a lot of attention from the vision and learning communities. This ambitious research field strives for the creation of intelligent agents that can interact with the surrounding environment. Smart interactions, however, require fine-grained perception and effective planning abilities. For this reason, current research focuses on the creation of rich and complex architectures that are trained in simulation with a large amount of data. Thanks to powerful simulating platforms~\cite{deitke2020robothor,savva2019habitat,xia2018gibson}, the Embodied AI community could achieve nearly perfect results on the PointGoal Navigation task (PointNav)~\cite{wijmans2019dd}. However, current research is still in the first mile of the race for the creation of intelligent and autonomous agents. Naturally, the next milestones involve bridging the gap between simulated platforms (in which the training takes place) and the real world~\cite{kadian2020sim2real}. In this work, we aim to design a robot that can navigate in unknown, real-world environments.

We ask ourselves a simple research question: \textit{can the agent transfer the skills acquired in simulation to a more realistic setting?} To answer this question, we devise a new experimental setup in which models learned in simulation are deployed on a LoCoBot~\cite{locobot}. Previous work on Sim2Real adaptability from the Habitat simulator~\cite{savva2019habitat} has focused on a setting where the real-world environment was matched with a corresponding simulated environment to test the Sim2Real metric gap. To that end, Kadian~\etal~\cite{kadian2020sim2real} carry on a 3D acquisition of the environment specifically built for robotic experiments. Here, we assume a setting in which the final user cannot count on the technology/expertise required to make a 3D scan. This experimental setup is more challenging for the agent, as it cannot count on semantic priors on the environment acquired in simulation. Moreover, while~\cite{kadian2020sim2real} employs large boxes as obstacles, our testing scene contains real-life objects with complicated shapes such as desks, office chairs, and doors.

\begin{figure}[t!]
    \centering
    \includegraphics[width=0.9\linewidth]{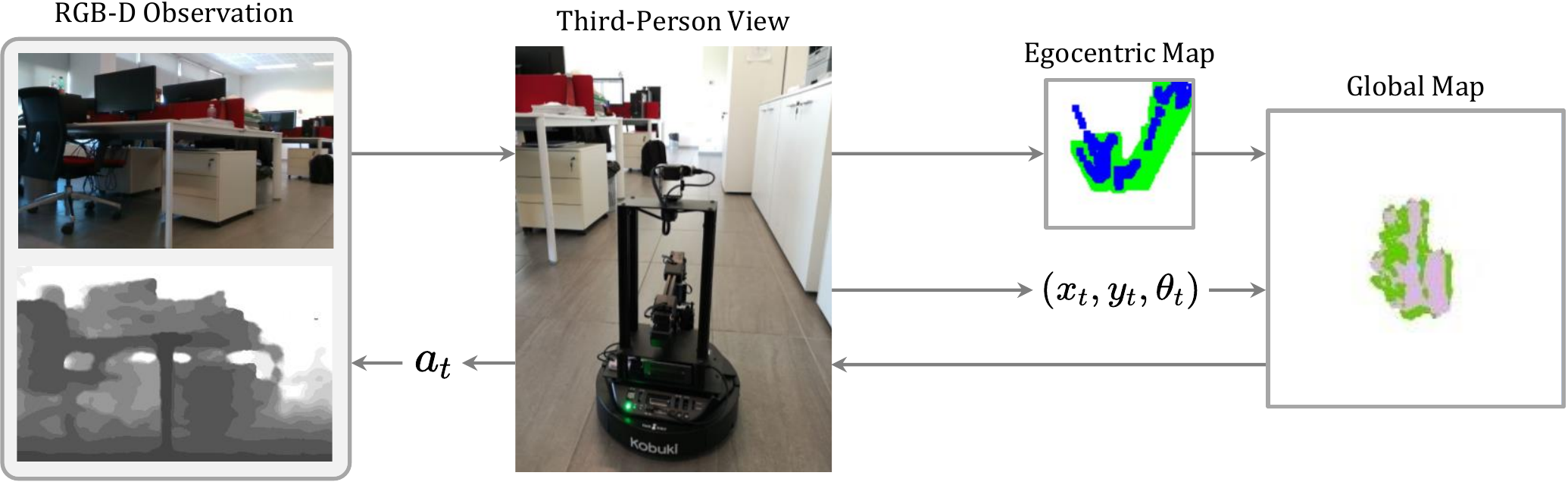}
    \caption{We deploy a state-of-art navigation architecture on a LoCoBot and test it in a realistic, office-like environment. Our model exploits egocentric and global occupancy maps to plan a route towards the goal.}
    \label{fig:fig1}
\end{figure}

Our agent builds on a recent model proposed by Ramakrishnan~\etal~\cite{ramakrishnan2020occupancy} for the PointNav task. As a first step, we research the optimal setup to train the agent in simulation. We find out that default options (tailored for simulated tasks) are not optimal for real-world deployment: for instance, the simulated agents often exploit imperfections in the simulator physics to slide along the walls. As a consequence, deployed agents tend to get stuck when trying to replicate the same sliding dynamic. By enforcing a more strict interaction with the environment, it is possible to avoid such shortcomings in the locomotor policy. Secondly, we employ the software library PyRobot~\cite{murali2019pyrobot} to create a transparent interface with the LoCoBot: thanks to PyRobot, the code used in simulation can be seamlessly deployed on the real-world agent by changing only a few lines of code. Finally, we test the navigation capabilities of the trained model on a real scene: we create a set of navigation episodes in which goals are defined using relative coordinates. While previous tests were mainly made in robot-friendly scenarios (often consisting of a single room), we test our model, which we call LoCoNav, in a more realistic environment with multiple rooms and typical office furniture~(Fig.\ref{fig:fig1}). Thanks to our experiments, we show that models trained in simulation can adapt to real unseen environments. By making our code and models publicly available, we hope to motivate further research on Sim2Real adaptability and deployment in the real world of agents trained on the Habitat simulator. 

\section{Related Work}
There is a broad area of recent research that focuses on designing autonomous agents with different abilities. Among these, a vast line of work concentrates on embodied exploration and navigation~\cite{chaplot2019learning,chen2019learning,ramakrishnan2020occupancy,ramakrishnan2020exploration}. In this setting, the agent's goal is to explore a new environment in the shortest amount of time. Architectures trained for this task usually employ reinforcement learning to maximize coverage (the area seen during a single episode)~\cite{chaplot2019learning}, surprisal~\cite{bigazzi2020explore}, or a reward based on the novelty of explored areas~\cite{ramakrishnan2020exploration}. Usually, this is done by creating internal map representations to keep track of the exploration progress and at the same time help the agent plan for future destinations~\cite{chaplot2019learning,chen2019learning,ramakrishnan2020occupancy}. The main advantage of these approaches is their ability to adapt to downstream tasks, such as PointGoal~\cite{ramakrishnan2020occupancy} or ObjectGoal~\cite{chaplot2020object} navigation. In PointGoal navigation, the target destination is specified using relative coordinates \textit{w.r.t.}~the agent's initial position and heading~\cite{savva2019habitat}. Using simulation and impressive computational power, Wijmans~\etal~\cite{wijmans2019dd} achieve nearly perfect results. However, their model is trained using $2.5$ billion frames and requires experience acquired over more than half a year of GPU time. Unfortunately, models tend to learn simulator-specific tricks to circumvent navigation difficulties~\cite{kadian2020sim2real}. Since such shortcuts do not work in the real world, there is a significant Sim2Real performance gap.

Recent work has studied how to deploy models trained on simulation to the real world~\cite{deitke2020robothor,kadian2020sim2real,rosano2020embodied}. In their work, Kadian \etal~\cite{kadian2020sim2real} make a 3D acquisition of a real-world scene and study the Sim2Real gap for various setups and metrics. However, their environment is very simple as obstacles are large boxes, the floor has an even and regular surface in order to facilitate the actuation system, and there are no doors or other navigation bottlenecks. In this work, instead, we focus on a more realistic type of environment: obstacles are represented by common office furniture such as desks, chairs, cupboards; the floor is uneven as there are gaps between floor tiles that make actuation noisy and very position-dependent, and there are multiple rooms that must be accessed through doorways.
\section{Real-World Navigation with Habitat}
In this section, we describe our out-of-the-box navigation robot. First, we describe the baseline architecture and its training procedure that takes place in the Habitat simulator~\cite{savva2019habitat}. Then we present our LoCoNav agent, which builds upon the baseline and implements various modules to enable real-world navigation.


\subsection{Baseline Architecture}
We draw inspiration from the occupancy anticipation agent~\cite{ramakrishnan2020occupancy} to design our baseline architecture. The model consists of three main parts: a mapper, a pose estimator, and a hierarchical policy, that we describe in the following.

\tinytit{Mapper}
The mapper is responsible for producing an occupancy map of the environment, which is then employed by the agent as an auxiliary representation during navigation. We use two different types of map at each time step $t$: the agent-centric map $v_t$ that depicts the portion of the environment immediately in front of the agent, and the global map $m_t$ that captures the area of the environment already visited by the agent. The global map of the environment $m_t$ is blank at $t=0$ and it is built in an incremental way. Each map has two channels, identifying the free/occupied and the explored/unexplored space, respectively; each pixel contains the state of a $5$cm $\times$ $5$cm area.
The mapper module takes as input the RGB and depth observations $(o^r_t, o^d_t)$ at time $t$ and produces the agent-centric map $v_t \in [0,1]^{2 \times V \times V }$. The RGB observation is encoded to a feature representation $\bar{o}^r_t$ with the first two layers of a pretrained ResNet-18 followed by a three-layered CNN. Instead, the depth observation is used to create a point-cloud and reprojected to form a preliminary map $\bar{o}^d_t$. The resulting agent-centric map $v_t$ is computed by combining $\bar{o}^r_t$ and $\bar{o}^d_t$ with a U-Net. Then, $v_t$ is registered to the global map $m_t \in [0,1]^{2 \times W \times W}$, with $W > V$, using the agent's position and heading in the environment $(x_t, y_t, \theta_t)$. 

\tinytit{Pose Estimator}
While the agent navigates towards the goal, the interactions with the environment are subject to noise and errors, so that, for instance, the action \textit{go forward $25$cm} might not result in a real displacement of $25$cm. That could happen for a variety of reasons: bumping into an obstacle, slipping on the terrain, or simple actuation noise. The pose estimator is responsible of avoiding such positioning mistakes and keeps track of the agent pose in the environment at each time step $t$. This module computes the relative displacement $(\Delta x_t, \Delta y_t, \Delta\theta_t)$ caused by the action selected by the agent at time $t$. It takes as input the RGB-D observations $(o^r_t, o^d_t)$ and $(o^r_{t-1}, o^d_{t-1})$ retrieved at time $t$ and $t-1$, and the egocentric maps $v_t$ and $v_{t-1}$. Each modality is considered separately to obtain a first estimate of the displacement:
\begin{equation}
    g_i = W_1 \text{max}(W_2\star + b_2,0) + b_1 ,
    \label{eq:g_i}
\end{equation}
The final output of the pose estimator is the weighted sum of the three displacement vectors $g_i$:
\begin{equation}
    (\Delta x_t,\Delta y_t,\Delta \theta_t) = \sum_{i=0}^{2}{\alpha_i \cdot g_i},  \qquad
    \alpha_i = \text{softmax}(\text{MLP}_i([\bar{o}^r_t, \bar{o}^d_t, \bar{v}_t])),
\end{equation}
where MLP is a three-layered fully-connected network, ($\bar{o}^r_t$, $\bar{o}^d_t$, $\bar{v}_t$) are the inputs encoded by a CNN and $[\cdot, \cdot, \cdot]$ denotes tensor concatenation. The estimated pose of the agent at time $t$ is given by 
$(x_t, y_t, \theta_t) = (x_{t-1}, y_{t-1}, \theta_{t-1}) + (\Delta x_t,\Delta y_t,\Delta \theta_t)$.

\tinytit{Hierarchical Policy}
Following a current trend in Embodied AI~\cite{chen2019learning,chaplot2019learning,ramakrishnan2020occupancy}, we employ a hierarchical policy in our baseline navigator. The highest-level component of our policy is the global policy. The global policy selects a long-term goal on the global map, that we call global goal. The input of the global policy at time $t$ is a $4$-channel enriched global map $m_t^+ \in [0,1]^{4 \times W \times W}$ obtained as the concatenation of the global map $m_t$ with a spatial representation of visited states and a one-hot representation of the agent position at time $t$. Finally, we compute an $8$-channel input of shape $G \times G$ for the global policy. To that end, we concatenate a cropped and a max-pooled version of $m_t^+$. The global policy outputs a probability distribution over the $G \times G$ action space. The global goal is sampled from this distribution and is then converted to $(x,y)$ coordinates on the global map. A new global goal is sampled every $N$ time steps during training and is set to the navigation goal during deployment and test.
The middle-level component of our hierarchical policy is the planner. After the global goal is set, an A* planner decodes the next local goal within $0.25$m from the agent and on the trajectory towards the global goal. A new local goal is sampled if at least one of the following three conditions verifies: a new global goal is sampled by the global policy, the previous local goal is reached, or the local goal is known to be in an occupied area.
Finally, the local policy performs the low-level navigation and decodes the series of actions to perform. The actions available to the agents are \textit{go forwards $25$cm} and \textit{turn $15$\textdegree}. The local policy samples an atomic action $a_t$ at each time step $t$.

\subsection{Training in Simulation}
The baseline architecture described in the previous lines is trained in simulation using Habitat~\cite{savva2019habitat} and 3D scans from the Gibson dataset of spaces~\cite{xia2018gibson}. The mapper is trained with a binary cross-entropy loss using the ground-truth occupancy maps of the environment, obtained as described in~\cite{ramakrishnan2020occupancy}. The navigation policy is trained using reinforcement learning. We choose PPO~\cite{schulman2017proximal} as training algorithm. The global policy receives a reward signal equal to the increase in terms of anticipated map accuracy~\cite{ramakrishnan2020occupancy}:
\begin{equation}
    R_t^{\text{\textit{glob}}} = \text{Accuracy}(m_t, \hat{m}) - \text{Accuracy}(m_{t-1}, \hat{m}) ,
\end{equation}
where $m_{t}$ and $m_{t-1}$ represent the global occupancy maps computed at time $t$ and $t-1$ respectively, and $\hat{m} \in [0,1]^{2 \times W \times W}$ is the ground-truth global map.
%
The map accuracy is defined as: 
\begin{equation}
    \text{Accuracy}(m,\hat{m}) = \sum_{i = 1}^{W^2}{\sum_{j=1}^{2}{\mathbbm{1}[m_{ij} = \hat{m}_{ij}]}},
\end{equation}
where $\mathbbm{1}[\cdot]$ is an indicator function that returns one if the condition $[\cdot]$ is true and zero otherwise.
The local policy is trained using a reward that encourages the decrease in the euclidean distance between the agent and the local goal while penalizing collisions with obstacles:
\begin{equation}
    r_t^{local} = d_t - d_{t-1} - \alpha * \text{\textit{bump}}_t,
    \label{eq:r_t^local}
\end{equation}
where $d_t$ and $d_{t-1}$ are the euclidean distances to the local goal at times $t$ and $t-1$, $\text{\textit{bump}}_t \in \{0,1\}$ identifies a collision at time $t$ and $\alpha$ regulates the contributions of the collision penalty.
The training procedure described in this section exploits the experience collected throughout $6.5$ million exploration frames.

\subsection{LoCoNav: Adapting for Real World}
\label{loconav}
The baseline architecture described above is trained in simulation and achieves state-of-art results on embodied exploration and navigation~\cite{ramakrishnan2020occupancy}. The reality, however, poses some major challenges that need to be addressed to achieve good real-world performances. For instance, uneven ground might give rise to errors and noise in the actuation phase. To overcome this and other discrepancies between simulated and real environments, we design LoCoNav: an agent that leverages the availability of powerful simulating platforms during training but is tailored for real-world use. In this section, we describe the main characteristics of the LoCoNav design.
We deploy our architecture on a LoCoBot~\cite{locobot} and use PyRobot~\cite{murali2019pyrobot} for seamless code integration.

\tinytit{Prevent your Agent from Learning Tricks}
All simulations are imperfect. One of the main objectives when training an agent for real-world use in simulation is to prevent it from learning simulator-specific tricks instead of the basic navigation skills. During training, we observed that the agent tends to hit the obstacles instead of avoiding them. This behavior is given by the fact that the simulator allows the agent to slide towards its direction even if it is in contact with an obstacle as if there were no friction at all. Unfortunately, this ideal situation does not fit the real world, as the agent needs to actively rotate and head towards a free direction every time it bumps into an obstacle. To replicate the realistic \textit{sticky} behavior of surfaces, we check the \textit{bump}$_t$ flag before every step. If a collision is detected, we prevent the agent from moving forward. As a result, our final agent is more cautious about any form of collision. 

\begin{table}[b!]
    \centering
    \setlength{\tabcolsep}{.4em}
    \caption{List of hyperparameters changes for Sim2Real transfer.}
    \label{tab:param}
    \resizebox{\linewidth}{!}{
    \begin{tabular}{lcccccc}
    \toprule
         & & \textbf{Height} & \textbf{RGB FoV} & \textbf{Depth FoV} & \textbf{Depth Range} & \textbf{Obst. Height Thresh.}\\
    \midrule
        \textbf{Default for Simulation} & & 1.25 & \textbf{H}: 90, \textbf{V}: 90 & \textbf{H}: 90, \textbf{V}: 90 & [0.0, 10.0] & [0.2, 1.5]\\
        \textbf{LoCoNav (ours)} & & 0.60 & \textbf{H}: 70, \textbf{V}: 90 & \textbf{H}: 57, \textbf{V}: 86 & [0.0, 5.00] & [0.3, 0.6]\\
    \bottomrule
    \end{tabular}
    }
\end{table}

\tinytit{Sensor and Actuation Noise}
Another important discrepancy between simulation and real-world is the difference in the sensor and actuation systems. Luckily, the Habitat simulator allows for great customization of input-output dynamics, thus being very convenient for our goal.
In order to train a model that is more resilient to the camera noise, we apply a Gaussian Noise Model on the RGB observations and a Redwood Noise Model~\cite{choi2015robust} on the depth observations. Unfortunately, the LoCoBot RealSense camera still presents various artifacts and regions with missing depth values. For that reason, we need to restore the observation retrieved from the depth camera before using it in our architecture. To that end, we apply the hole filling algorithm described in~\cite{telea2004image}, followed by the application of a median filter.

Regarding the actuation noise, we find out that the use of the incremental pose estimator (employed in the occupancy anticipation model and described in our baseline architecture) is not optimal, especially when combined with the actuation noise typical of real-world scenarios. Luckily, we can count on more precise and reliable information coming from the LoCoBot actuation system. By checking the actual rotation of each wheel at every time step, the robot can update its position step by step. We adapt the odometry sensor of the LoCoBot platform to be compliant with our architecture. To that end, the pose returned by the sensor is converted by resetting it with respect to its state at the beginning of the episode. We name $\mathbf{\chi}_0 = (x_0, y_0, \theta_0)$ the coordinate triplet given by the odometry sensor at $t=0$. We then define:
\begin{equation}
    \mathbf{A} = \begin{pmatrix}
            \mathbf{R}_0 & \mathbf{t}_0\\
            \mathbf{0} & 1 
            \end{pmatrix}
            = 
            \begin{pmatrix}
            \cos{\theta_0} & -\sin{\theta_0} & x_0\\
            \sin{\theta_0} & \cos{\theta_0} & y_0\\
            0 & 0 & 1 
            \end{pmatrix}.
\end{equation}
Let us define $\mathbf{x}_t$ as the augmented position vector $(x_t, y_t, 1)$ containing the agent position at each step $t$. We compute the relative position of the robot as:
\begin{equation}
    \mathbf{\tilde{x}}_t = \mathbf{A}^{-1}\mathbf{x}_t, \qquad \tilde{\theta}_t = \theta_t - \theta_0
\end{equation}
where $\mathbf{\tilde{x}}_t = (\tilde{x}_t, \tilde{y}_t, 1)$ contains the agent position after the conversion to episode coordinates. The relative position and heading is given by $\mathbf{\tilde{\chi}}_t = (\tilde{x}_t, \tilde{y}_t, \tilde{\theta}_t)$.
Note that, for $t=0$, $\mathbf{\tilde{\chi}}_0 = (\tilde{x}_0, \tilde{y}_0, \tilde{\theta}_0) = (0,0,0)$.

\tinytit{Hyperparameters}
Finally, we noticed that typical hyperparameters employed in simulation do not match the real robot characteristics. For instance, the camera height is set to $1.25$m in previous works, but the RealSense camera on the LoCoBot is placed only $0.6$m from the floor. During the adaptation to the real-world robot, we change some hyperparameters to align the observation characteristics of the simulated and the real world and to match real robot constraints. These parameters are listed in Table \ref{tab:param}.
\section{Experiments}
\begin{figure}[t!]
    \begin{minipage}[b][][b]{0.54\textwidth}
        \centering
        \footnotesize
        \includegraphics[width=0.9\linewidth]{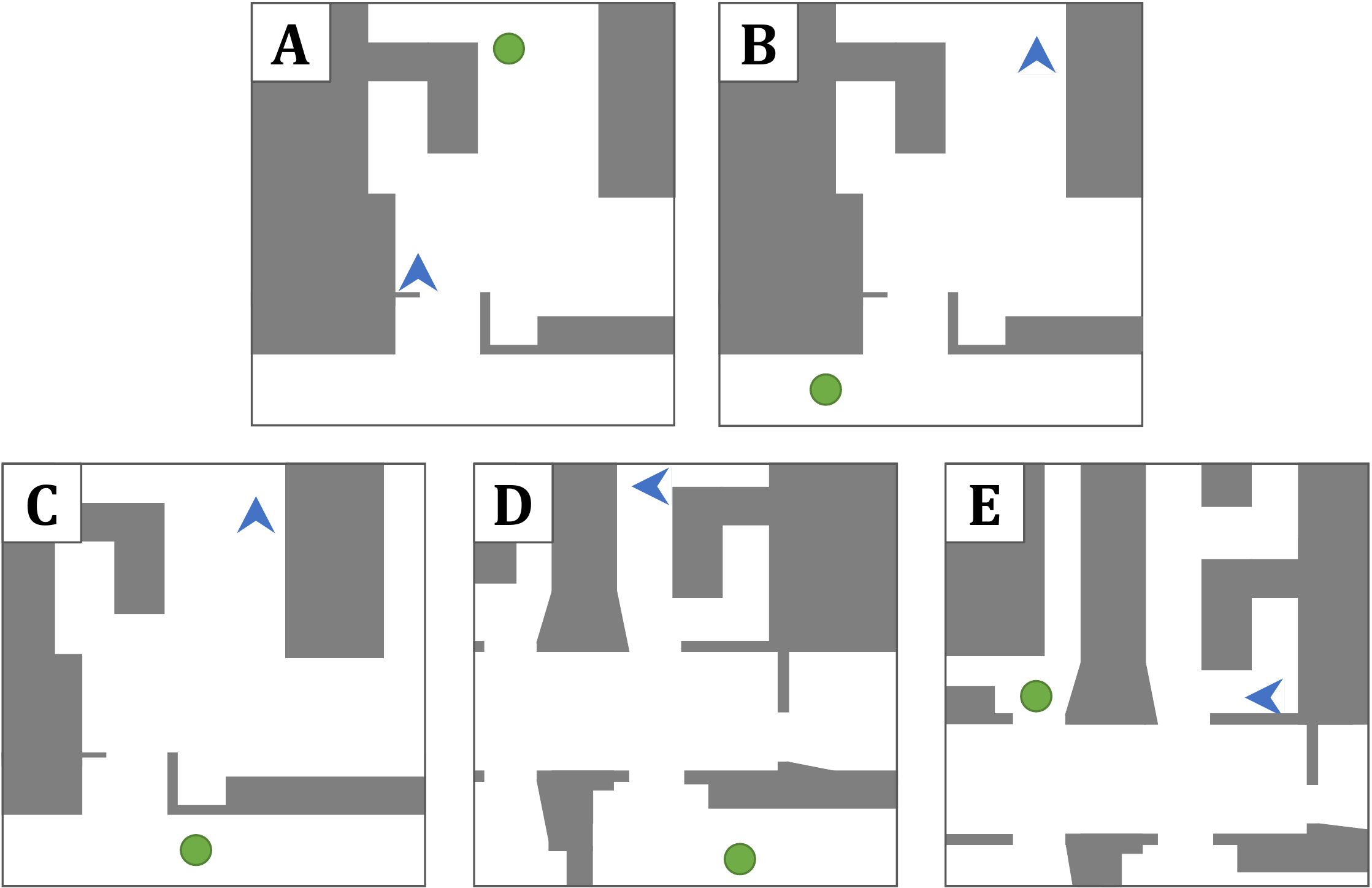}\\
        (a)
    \end{minipage}
    \hspace{0.3cm}
    \begin{minipage}[b][][b]{0.42\textwidth}
        \centering
        \footnotesize
        \resizebox{0.98\linewidth}{!}{
\setlength{\tabcolsep}{.25em}
\begin{tabular}{cc ccc}
\toprule          
\textbf{Path} & & \textbf{Length(m)} & \textbf{Time(s)} & \# \textbf{Step} \\
\midrule
\textbf{A} & & 3.80 & 124 & 23  \\
\textbf{B} & & 6.75 & 239 & 45  \\
\textbf{C} & & 5.95 & 223 & 43  \\
\textbf{D} & & 6.55 & 217 & 42  \\
\textbf{E} & & 4.20 & 227 & 33  \\
\bottomrule
\end{tabular}
}\\
        \vspace{0.1cm}
        (b)
    \end{minipage}
    \caption{Layout of the navigation episodes (a). Path-specific information, as obtained with human supervision (b).}
    \label{fig:paths}
    \vspace{-0.45cm}
\end{figure}

\tinytit{Testing Setup} We run multiple episodes in the real environment, in which the agent needs to navigate from a starting point A to a destination B. The goal is specified by using relative coordinates (in meters) with respect to the agent's starting position and heading. Although the agent knows the position of its destination, it has no prior knowledge of the surrounding environment. Because of this, it cannot immediately plan a direct route to the goal and must check for obstacles and walls before stepping ahead. After each run, we reset the agent memory so that it cannot retain any information from previous episodes.
We design five different navigation episodes that take place in three different office rooms and the corridor connecting them (Fig.~\ref{fig:paths}a). For each episode, we run different trials with different configurations: obstacles are added/moved, or people are sitting/standing in the room. In total, we run $50$ different experiments, resulting in more than $10$ hours of real-world testing.

\tinytit{Evaluation Protocol} An episode is considered successful if the agent sends a specific \textit{stop} signal within $0.2m$ from the goal. This threshold corresponds to the radius of the robot base. For every navigation episode, we also track the number of steps and the time required to reach the goal. Since the absolute number of steps is not comparable among different episodes, we ask human users to control the LoCoBot and complete each navigation path via a remote interface (we report human performance in Fig~\ref{fig:paths}b). We then normalize these measures using this information so that results close to $1.00$ indicate human-like performances. We provide absolute and normalized length and time for each episode, as well as the popular SPL metric (Success rate weighted by inverse Path Length). We employ a slightly modified version of the SPL, in which the normalization is made basing on the number of steps and not on the effective path length to penalize purposeless rotations. Additionally, we set a boolean flag for each episode that signals whether the robot has bumped into an obstacle, and we report the average Bump Rate (BR). We also report the Hard Failure Rate (HFR) as the fraction of episodes terminated for one of the following reasons: the agent gets stuck and cannot proceed, or the episode length exceeds the limit of $300$ steps.

\begin{figure}[t!]
    \centering
    \includegraphics[width=0.9\linewidth]{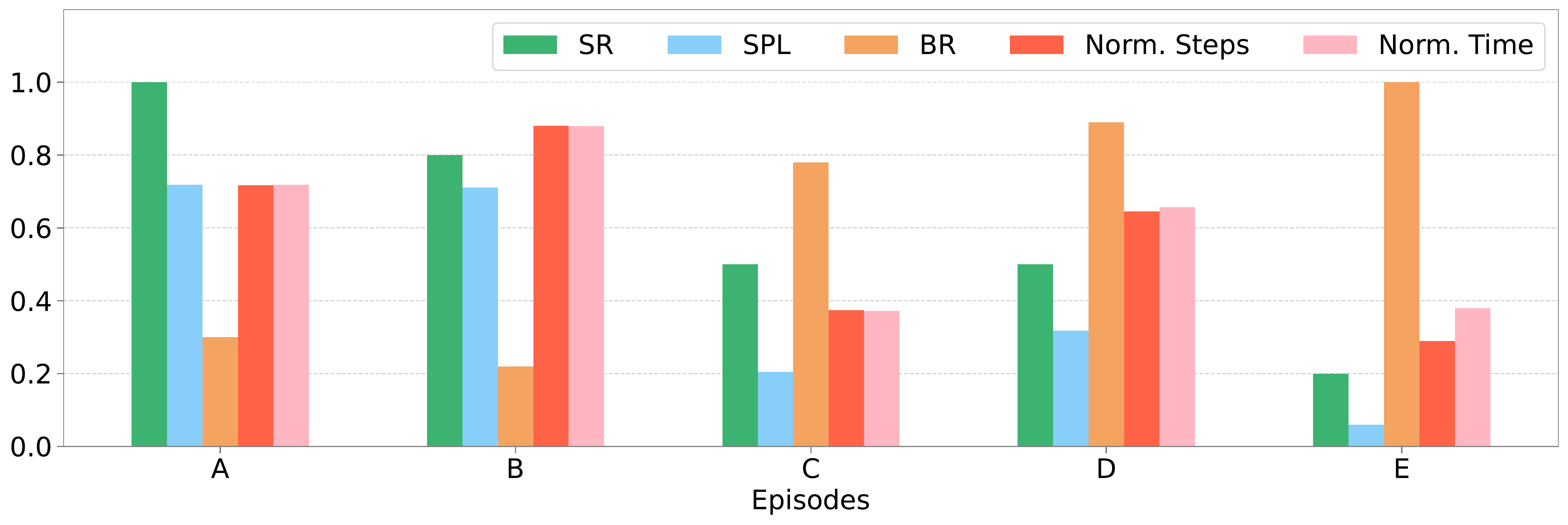}
    \vspace{-0.15cm}
    \caption{Comparison of the main navigation metrics on different episodes.}
    \label{fig:hists}
    \vspace{-0.5cm}
\end{figure}

\tinytit{Real-world Navigation}
In this experiment, we test our robot on five different realistic navigation paths (Fig.~\ref{fig:paths}a). We report the numerical results for these experiments in Table~\ref{tab:results}, and we plot the main metrics in Fig.~\ref{fig:hists} to allow for a better visualization of navigation results across different episodes. When a path is contained in a single room (A), the agent achieves optimal results, as it always stops within the success threshold from the goal. The number of steps is slightly higher than the minimum required by the episode ($33$ instead of $23$), but this overhead is necessary as the agent must rotate and ``look around'' to build a decent map of the surrounding before planning a route to the goal. Paths that involve going outside the room and navigating different spaces (B, C, D, E) are fairly complicated, but the agent can generally terminate the episode without hard failures. When the shortest path to the goal leads to a wall or a dead-end, the agent needs to find an alternative way to circumvent this obstacle (\eg~a door). This leads to a higher episode length because the robot must dedicate some time to general exploration of the surroundings. Finally, we find out that the most challenging scenario for our LoCoNav is when reaching the goal requires to get out of a room and then enter a door immediately after, on the same side of the corridor (as in E). Since the robot sticks to the shortest path, the low parallax prevents it from identifying the second door correctly. Even in these cases, a bit of general exploration helps to solve the problem.
%
% ...
\begin{table}[t]
\centering
\caption{Navigation results. Numbers after $\pm$ denote the standard error of the mean.}
\label{tab:results}
\setlength{\tabcolsep}{.5em}
\resizebox{0.98\linewidth}{!}{
\begin{tabular}{cc cccc cc cc }
\toprule          
\textbf{Path} & & \textbf{SR} $\uparrow$ & \textbf{SPL} $\uparrow$ & \textbf{HFR} $\downarrow$ & \textbf{BR} $\downarrow$
& \textbf{Abs.~Steps} & \textbf{Norm.~Steps} $\uparrow$ & \textbf{Abs.~Time} & \textbf{Norm.~Time} $\uparrow$ \\
\midrule
\textbf{A} & & 1.0 & 0.718 &  0.0 & 0.30 & 32.70$\pm$1.73 & 0.717$\pm$0.033 & 176.11$\pm$10.39 & 0.718$\pm$0.031 \\
\textbf{B} & & 0.8 & 0.711 & 0.10 & 0.22 & 51.67$\pm$1.72 & 0.880$\pm$0.027 & 273.70$\pm$8.24 & 0.879$\pm$0.030 \\
\textbf{C} & & 0.5 & 0.205 & 0.10 & 0.78 & 123.44$\pm$10.66 & 0.374$\pm$0.034 & 631.15$\pm$50.09 & 0.372$\pm$0.036 \\
\textbf{D} & & 0.5 & 0.318 & 0.10 & 0.89 & 65.67$\pm$3.90 & 0.645$\pm$0.037 & 344.00$\pm$20.08 & 0.657$\pm$0.038 \\
\textbf{E} & & 0.2 & 0.060 & 0.40 & 1.00 & 135.17$\pm$29.97 & 0.290$\pm$0.049 & 722.76$\pm$162.01 & 0.38$\pm$0.066 \\
\midrule
\textbf{Overall} & & 0.6 & 0.402 & 0.14 & 0.60 & \textbf{-} & 0.608$\pm$0.036 & \textbf{-} & 0.617$\pm$0.034 \\
\bottomrule
\end{tabular}
}
\end{table}

\tinytit{Discussion and Failure Cases}
Overall, our experimental setup provides a challenging test-bed for real-world robots. We find out that failures are due to two main issues. First, when the agent must navigate to a different room, it has no access to a map representing the general layout of the environment. This prevents the robot from computing a general plan to reach the long-term goal and forces it to explore the environment before proceeding. If a map was given to the agent, this problem would have been greatly alleviated. A second problem arises when the goal is close in terms $(x,y)$ coordinates but is physically placed in an adjacent room. To solve this problem, one could decompose the navigation between rooms in a multi-goal problem where neighboring nodes are closer. In this way, it is possible to reduce a complex navigation episode in simpler sub-episodes (like A or B), in which our agent has proved to be successful. 
\section{Conclusion}
We have presented LoCoNav, an out-of-the-box architecture for embodied navigation in the real world. Our model takes advantage of two main elements: state-of-art simulating platforms, together with a large number of 3D spaces, for efficient and fast training, and a series of techniques specifically designed for real-world deployment. Experiments are conducted in reality on challenging navigation paths and in a realistic office-like environment. Results demonstrate the validity of our approach and encourage further research in this direction.

\footnotesize{\section*{Acknowledgment}
This work has been supported by ``Fondazione di Modena'' under the project 
``AI for Digital Humanities'' and by the national project ``IDEHA: Innovation for Data Elaboration in Heritage Areas'' (PON ARS01\_00421), cofunded by the Italian Ministry of University and Research.}

\bibliographystyle{splncs04}
\bibliography{references}

\end{document}